\documentclass[11pt,a4paper]{article}
\usepackage[hyperref]{acl2017}
\usepackage{times}
\usepackage{latexsym}
\usepackage{float}
\usepackage{graphicx}
\usepackage{url}
\usepackage{datetime}

\aclfinalcopy % Uncomment this line for the final submission
\title{ Entity linking with people entities on Wikipedia}
\author{Weiqian Yan \and Kanchan Khurad \\
  Department of Math and Computer Science Emory University \\
  { \tt weiqian.yan@emory.edu} \and \tt kkhurad@emory.edu}

\begin{document}
\maketitle
\begin{abstract}
This paper introduces a new model that uses named entity recognition,
coreference resolution, and entity linking techniques, to approach the
task of linking people entities on Wikipedia people pages to their
corresponding Wikipedia pages if applicable. Our task is different
from general and traditional entity linking because we are working in
a limited domain, namely, people entities, and we are including
pronouns as entities, whereas in the past, pronouns were never
considered as entities in entity linking. We have built 2 models, both
outperforms our baseline model significantly. The purpose of our
project is to build a model that could be use to generate cleaner data
for future entity linking tasks. Our contribution include a clean data
set consisting of 50 Wikipedia people pages, and 2 entity linking
models, specifically tuned for this domain.
\end{abstract}
\section{Introduction}
Entity linking is a developing and interesting field in natural
language processing research.It has been used in improving the performance of information retrieval system, information summarization, and many such other applications.One of the  issues in current entity linking
research is the lack of data sets, and the data sets that are available are noisy, so as to contribute towards this shortcoming, we have built a model which could be used to generate training data set for entity linking tasks. The major difference between the conventional entity linking systems and our model is that, our model not only recognizes the entities in a text but also recognizes the pronouns referring to those entities. We have manually annotated Wikipedia pages to evaluate the performance of our model. We have restricted our experiment only to Wikipedia pages about people, and our model recognizes only the entities addressing a particular person, the entities include pronouns as well as proper nouns. Given a Wikipedia page on people, our model will find person
name entities, and pronouns and link them to their corresponding
Wikipedia page. We have worked on data of a single domain (people) and
we have achieved pretty good results using our model.We have used F1
score to evaluate the performance of our model as well as to evaluate
the performance of the components of our model, which are Named Entity
Recognition, coreference resolution.We have used Stanford's Named
entity recognition and Stanford's coreference resolution in our
experiment.\\
We will briefly summarize the related works to each component in our
model in section 2. In section 3, we will thoroughly explain our data,
since it is an important contribution in our project. We will also
explain our experiment settings. We will present our baseline model,
and our optimized models in detail. We include diagrams for both to
give a better and visual understanding of our models. We provide
analysis on our model, based on the performance of our model on
different type of samples in our test data set. Section 4 is the
conclusion we have reached from this project. Section 5 proposes
possible future works that could potentially improve the performance of the
models presented in this paper. 
\section{Related Work}
 To the best of our knowledge, this is the first time that this task
 has been attempted. However, there are related works for each
 component in our model,namely,named entity recognition, coreference
 resolution, and TagMe entity linking model. 
\begin{itemize}
\item We used Stanford's named entity recognition in the mention
  detection phase of our model. Stanford's named entity recognition is
  a linear chain Conditional Random Field model. The model is similar
  to the baseline local+Viterbi model in \cite{CRF}, but with added
  distributional similarity based features. We use Stanford's NER to
  extract all people's names in a Wikipedia page, but since we are
  working on a limited domain, we have added ruled based features to
  the NER model.
\item We used Stanford's coreference resolution in the mention
  detection phase of our model, in order to link pronouns to the names
  that they are referring to. Stanford's coreference resolution
  implements the multi-pass sieve coreference resolution system based
  on \cite{Lee} and \cite{Rag}. We also added rule based features to
  the coreference resolution system, in order to pick up pronouns that
  have been missed by the system. 
\item We used TagMe entity linking model in our baseline model, and
  one of our models. TagMe is an entity linking model, originally
  built for short texts, with Wikipedia as knowledge base. It is
  introduced in \cite{tagme} . 
\item We used Wikipedia search API for the entity linking in one of
  our models. Wikipedia search API uses Elastic Search, which is a
  search engine server on Lucene. It is a highly scalable, distributed
  and full-text search engine. It advances in speed, security,
  scalability, and hardware efficientcy, thus making it a very popular
  choice among enterprise search engines\cite{ES}.
\end{itemize}
\section{Experiment}
We have conducted experiments for each individual component in our
system, and our integrated systems.We think it is important to
evaluate the performance of each individual component in our models,
so that we could identify bottlenecks, and propose future
improvements. And we record recall, precision,
F-1 scores for all the experiments in Results. 
\subsection{Data}
Since it is the first time that anyone has attempted this specific
task, we were not able to find any existing data that we could use to evaluate our system. Hence we
manually annotated 50 Wikipedia people pages, including people in
sports, chefs, scientist, nobles and etc..The structure of our evaluation data set consists of four columns. In the first column for each of the Wikipedia
page in our test data set, we tokenized the texts, so that each line of fist column
contains one word/symbol, each row of the spreadsheet represents a word or a symbol and in the following columns we have identified the nature of the token, i.e if its an entity or not .In the second column we identify if the word is a name of a person with a label 'Y' (proper noun), in the third column we identify with a label 'Y' if the word in the first column is a name of a person and also identify if it is a pronoun referring a person.
The last column has the Wikipedia tag, if there exists a wikipedia page for the identified entities (proper nouns and pronouns both).The purpose behind having four columns was to evaluate each component of our model separately, with the annotations in the second column we evaluated the Named Entity Recognition system used in our model, with the annotations in the third column we could evaluate working of combined coreference resolution system and Named entity recognition system in our model and finally with the annotations in the fourth column we could evaluate the entity linking of our system . Some observations noted while annotating data were that, sometimes names of people can be used in various television shows for example "The Oprah Winfrey Show" has the name of a person who has a Wikipedia page but in our annotations we have not recognized it as an entity because it is a name of a show and as a whole it does not count as a name of person similarly some buildings or institutions which are named after people were ignored in the annotation. There were also some pages where awards were named after people, in this case as well we did not identify it as an entity .We annotated 50 Wikipedia pages separately for
evaluation purpose. The F1 score of our annotated test data is 97.2.\\

\begin{figure*}[t!]
  \includegraphics[width=\textwidth,height=8cm]{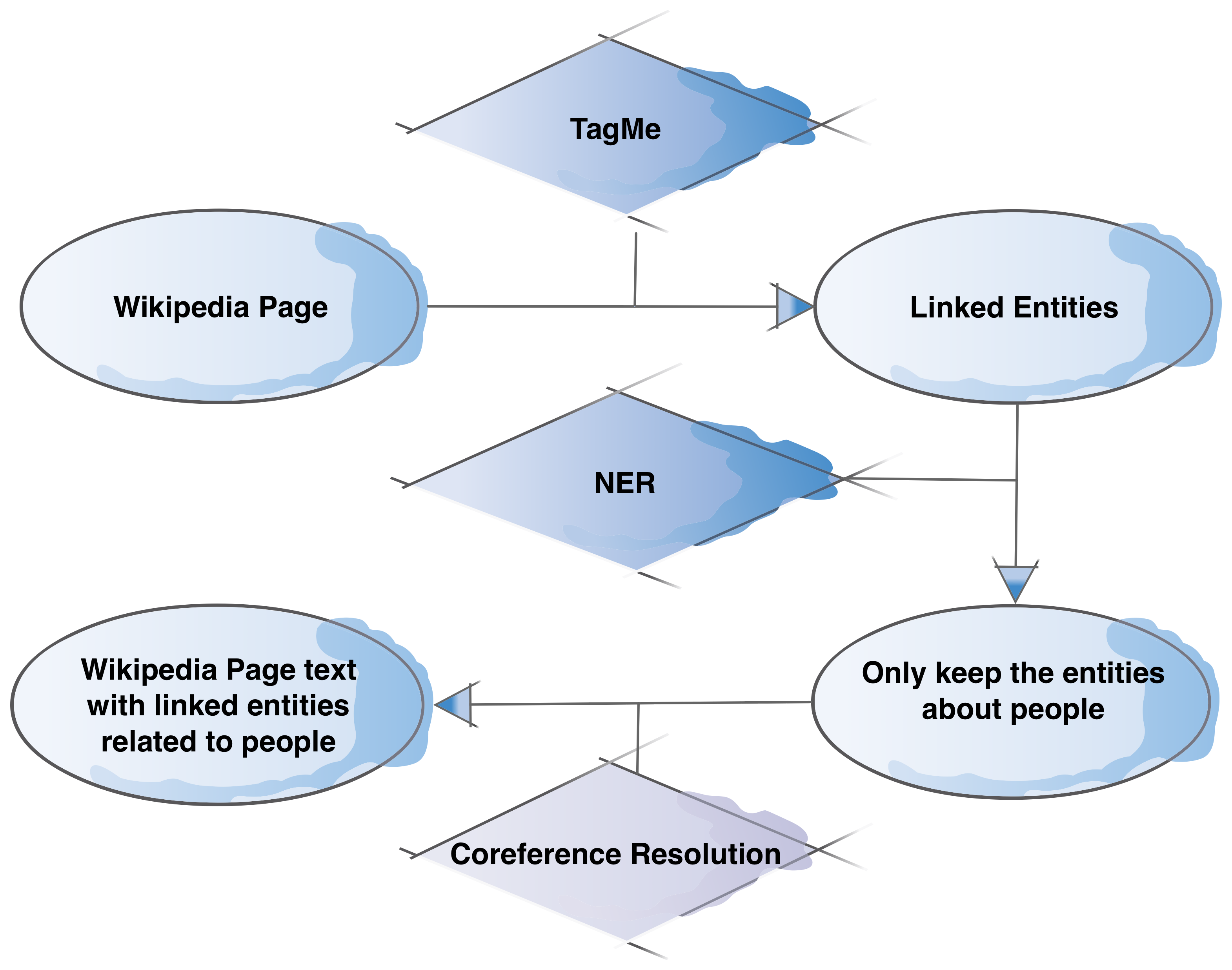}
  \caption{Baseline model.}
\end{figure*}

We also have an XML file for each of the Wikipedia page in our test
data set. The XML files has the information provided in the
information box for each person by Wikipedia. They contain information
like a person's first name, last name, gender, profession and
etc..this information we have used in the rule based features, which
we would elaborate in the next section. We have chosen to use XML to
record these personal information because XML is good at keeping
structural data, and easy to parse in Java.
\subsection{Approach}
In this section, we will introduce the baseline model we used and
models that we created, the baseline model uses TagMe and the other
model uses Wikipedia search or TagMe. We compare the performance of
our model with the baseline model, and report the score in Results section. 
\subsubsection{Baseline}
The baseline model that we used is a combination of named entity
recognition, coreference resolution and TagMe. NER is used to only
keep the people entities returned by TagMe, TagMe is a general entity
linking model with Wikipedia as knowledge base. So it would mark all
entities it has detected and link them to Wikipedia pages. We use NER only to 
keep the people entities, otherwise if we would have used NER to keep all the entities in our model then the recall would have been unnecessarily and
incorrectly low. We use coreference resolution, only to keep
pronouns and discard the non pronouns detected by the coreference system. Standford's coreference resolution is also a general model, it
marks all expressions,phrases and pronouns referring to a person. For example, on the Andre Trollope
Wikipedia page, the first, 'Sir Andrew Trollope (died 1461) was an
English soldier during the later stages of the Hundred Years' War and
at the time of the Wars of the Roses.', Stanford's coreference
resolution would link the phrase "English soldier during the later stages of
the HUndred Years Way" to Sir Andrew Trollope. But since the phrase is
not a pronoun, we won't keep it as a potential mention. TagMe is used
for entity linking in the baseline model. TagMe is initially built for
short texts, but some Wikipedia pages are long. So we set a limit. If
the Wikipedia page exceeds the length limit, we will process the
Wikipedia page sentence by sentence through TagMe, otherwise, we will
put the whole Wikipedia page through TagMe. Even though TagMe is built
for short texts, we found that using longer texts improves the
accuracy of entity linking in TagMe. We think this is due to the fact
that, with longer texts input, more contexts are provided for TagMe to
disambiguate potential entities. The flow of the model
is shown in figure 1. \\
\\
We have built 2 models in this project. The mention detection part for
both models is the same, we use named entity recognition, coreference
resolution  for the mention detection part in both  baseline and our
model, but we have used rule based features for optimization in out
models. The difference between our two models is that one uses Wikipedia
search API for entity linking, and the other uses TagMe for entity linking. The
general flow of our models is shown in the next page.
\begin{figure*}[h]
\includegraphics[width=\textwidth,height=8cm]{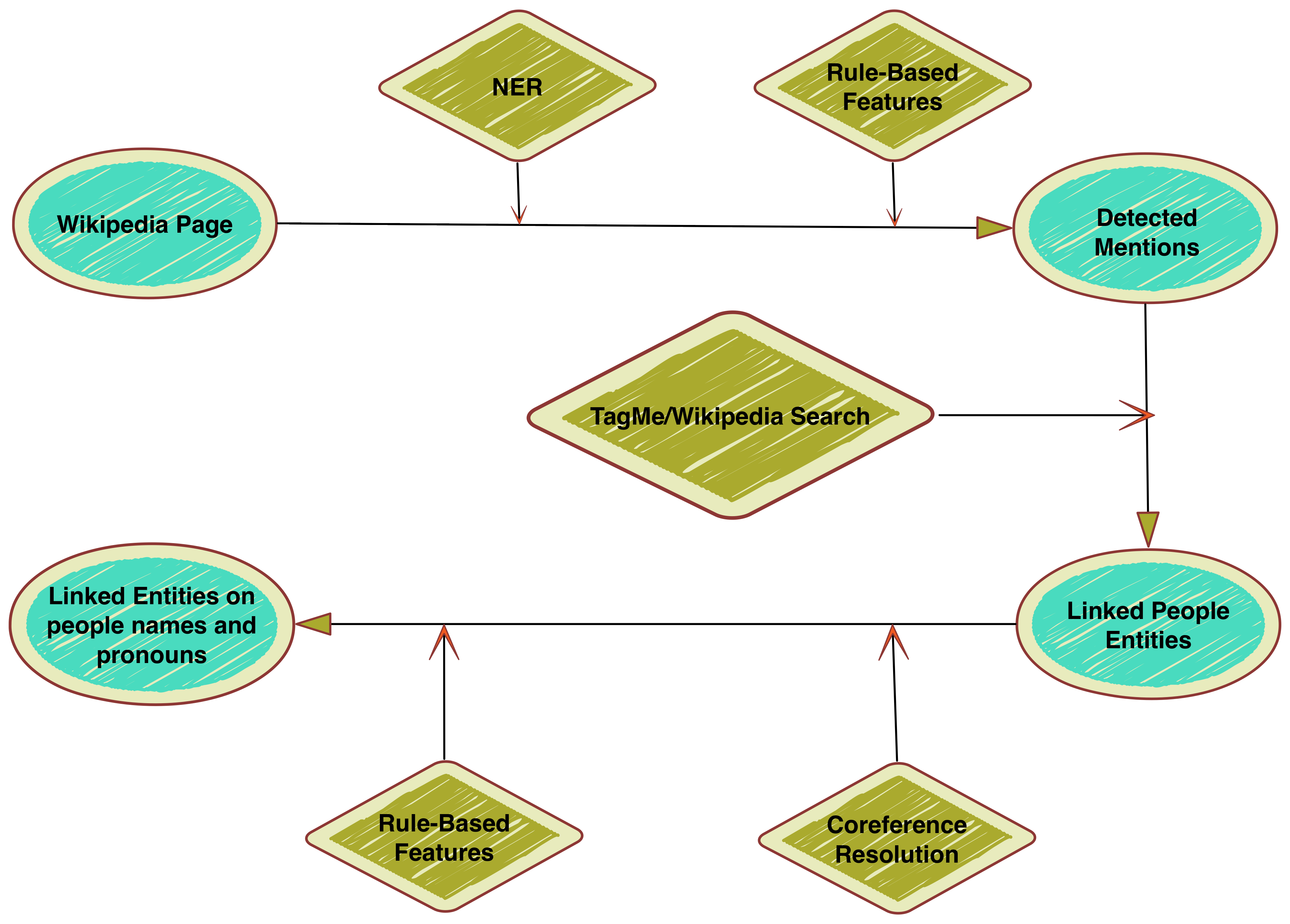}
\caption{System.}
\end{figure*}\\
We use named entity recognition to extract names of people from a
Wikipedia page. Then since we are limiting our domain  only to people,
we can get last name, first name, middle name if applicable,
gender and if available profession from the information box provided in the Wikipedia page. Using this information in form of XML files we added some
rule based features to the model after running the Wikipedia page through NER. We run through the
Wikipedia page again, to see if the model finds any token that matches 
with the person's first name or last name or middle name, and it marks that
token as a potential mention of this person. Also, we created a list of
titles e.g Sir, Lord, Miss, Dr. etc. in English. If the model finds that the first letter of the token is capitalized, and the token is in the list of titles, and it is
followed by a person's name, then it identifies the title to be a part of the person's 
name. Additionally, we noticed that sometimes people are referred as
name + location, especially nobles in ancient Europe. For example,
Margarate of Anjou is a French queen. Her name was Margarate, and she
was from Anjou, which is a city in France. The Stanford's NER is only
able to pick up Margarate as a potential mention, but if we only use
Margarate to link this mention, there could be thousands of people
with first name Margarate in Wikipedia, so it is necessary to include
the location as well. NER will mark location entities. So if we found
a person entity followed by of, and then a location entity, the model
will mark these as potential mention.\\
For the TagMe model, we will run each sentence that has at least one
potential mention through TagMe API, and record the labels it
returns. With NER, we only keep the labels for the entities that are
recognized as person or link to a person. \\
For the Wikipedia search model, we run each of the potential mention
through Wikiepdia search API, and if it returns more than 1 hits, we
will record the top one's label as the label for that mention. If a
mention consists of more than one token, we will put the tokens
together, separated by space, and then put them into Wikipedia search
API. In this part, since we are only using the top 1 hit returned by
Wikipedia,
we rely on Wikipedia's search algorithm, which is
Elastic Search. Elastic search can be used for full-text search. It
provides scalable searching. And it has near real time performance,
and can support multitenancy. \\
We use Stanford's coreference resolution in our model, to link
pronouns to the people that they are referring to. After coreference
resolution, we have added rule based features to the model. As
mentioned previously, we also only keep pronouns marked by Standford's
coreference resolution in this project. Since we
could possibly get gender from Wikipedia, and then we would run through the
document again, and if the model finds a unmarked pronoun that
matches this person's gender, it will link the token to this
person. For example, if the person is male, and the model found
unlinked pronouns in [he,him,his,himself], then the model would link
this mention with the person. The same logic applied to females, but
the pronouns list is [she,her,herself]. We also include [I,my,myself]
in the pronouns, as some Wikipedia pages contain quatations, which has
pronouns such as I referring to the person. We did't include plural
pronouns, like our, them, in our project, because they cause further
complication in coreference resolution. And it is impractical to link one
entity to multiple Wikipedia pages.
\subsection{Results}
Our model is an optimized version of the baseline model for the
mention detection part. We use the exactly same components, with rule
based features in our models. We have
chosen a limited domain, so we could make some assumptions on the
data, which we used to build the rule based features. For the entity
linking, we experimented with 2 different entity linking mechanism.  We have done
experiments on each individual component of our model vs. the baseline
model, and also our integrated model vs. the baseline model. For all
experiments, we record recall, precision, and F-1 scores.  Results
are shown in the table below:\\
\\
\begin{tabular}{ |p{1.55cm}|p{0.5cm}|p{1cm}|p{1.5cm}|p{1cm}| }
\hline
    &  & NER & Mention Detection & Entity Linking\\ \hline\hline
Baseline model & P & 89.6 & 92.2 & 51.7\\ 
			   & R & 93.3 & 82.2 & 48.6 \\ 
               & F & 90.8 & 86.6 & 49.5 \\ \hline\hline
Model (Wikipedia search) & P & 89.2 & 92.8 & 80.5\\ 
					    & R & 95.3 & 97.0 & 85.9\\ 
                        & F & 91.6 & 94.7 & 82.7\\ \hline\hline
Model (TagMe) & P & N/A & N/A & 71.9 \\
			 & R & N/A & N/A & 80.3\\
             & F & N/A & N/A & 75.5 \\ \hline
\end{tabular}\\
\\
In the above table R stands for Recall, P stands for Precision and F stands for F1 score.
As we can see from the experiment results, our model has increased
performance significantly over the baseline model. System with
Wikipedia search as the entity linked has the best performance,
reaching 82.7 in F-1 score. The results we got for both our models
have achieved significantly better results than the baseline model, in
precision, recall and F-1 score. NER in our model has good scores in
both recall and precision, which means the model is able to pick up
most of the people names mentioned in the Wikipedia page. Mention
detection is basically People names and Pronouns. The recall and
precision scores for mention detection is very good. So our model is
able to find the majority of potential mentions. For entity linking
with Wikipedia search, our best performing model, recall is good while
precision is not as good. We will discuss possible ways to improve the
entity linking in Future Work section.  
\subsection{Analysis}
We have analyzed the performance of our model on each sample of the
test data set, and found the following:\\
\begin{itemize}
\item The model has, in general, better performance in modern people
  than ancient people. We think the reason is: In the past, there are 
  very few first names available, so there
  are a lot of duplicate names. The only way to distinguish between
  these names are using the titles, for example, Richard Woodville,
  first Earl of Rivers. It's only possible to find the correct
  Wikipedia page for this person if his title is included. Including
  the title complicates the mention detection. There are multiple
  cases to consider for including titles. \\
\begin{itemize}
\item Name+of+Place: For example, Margarate of
  Anjou, where Anjou is a city in France. If we want to find the
  correct Wikipedia page using Wikipedia search, we have to input
  Margarate of Anjou, Margarate alone won't work, as there are
  so many people on Wikipedia with first name Margarate. Hence to find the correct Margarate in Wikipedia adding the place is essential, likewise for many other historical figures, it is necessary to have the  place in input along with the name.
\item Name+,+Title: In case of nobles, sometimes, they were referred by their
  full name followed by their title, like Richard Woodville, first
  Earl of Rivers. In noble families, father and son used to have the same
  name, the only difference was in the titles. For example, if the father is
 nth Earl of Warwick, and the son is (n+1)th Earl of
  Warwick, the identity of the father and son is only distinguished by the number (n)th. Without titles the identity of father and son would be the same and there will be no way to distinguish one from another, hence it is necessary to get the title of a historical figure, if it exists.
\item Title+Name: Noble people could also be referred to as their
  title followed by their first name, like King Edward VI. In this
  case, it's usually important to include the suffix, because there
  might be many people with the same title and first name, so that the
  suffix is the only way to distinguish them, like King Edward VI and
  his son King Edward V. 
\item If we consider people from further past, like  ancient Roman people, the
  entity disambiguation for the people in the ancient Rome, becomes even harder, because back in ancient Roman empire, there
  were only few first names available. Without enough historical
  background, even humans may find it difficult to identify these
  entities. For example, Gaius Julius Caesar is the full name for
  Caesar. Gaius can be a name, and it can also be used as a title,
  like Gaius Caeser, which is a different person from Gaius Julius
  Caesar. 
\end{itemize}
\item There are also some complications in modern names of people. There
 are some common English first and last names. If multiple
  people have the same first and last name, there are a few ways that
  Wikipedia uses to distinguish them.\\
\begin{itemize}
\item If the people have different middle name, Wikipedia will include
  their middle name in the label to distinguish them. So it's
  necessary for our model to also include middle name in searches. 
\item If middle name is not available, Wikipedia would use profession
  to distinguish these people. For example, Katie Cook and Katie
  Cook(writer). Sometimes, the personal info box provided by Wikipedia
  contains profession information, which we could use in our
  model. When profession is not available in the info box, entity
  linking becomes more complicated. 
\item Some people, usually actors/actresses, have stage name, which is
  different from their real name, like the real name of Vin Diesel is Mark
  Sinclair. In this case, it is important to use both the stage name
  and the real name for mention detection and entity linking. 
\end{itemize} 
\item Sometimes, Wikipedia will use nicknames or abbreviations to
  refer to people. For example, Alex for Alexander, Liz for Elizabeth
  and etc. In the Wikipedia page for Alex Guarnaschelli, his true
  first name is Alexandra, but through the Wikipedia page, he is
  referred to as Alex for most occurrences. 
\end{itemize}
Based on our observation, on Wikipedia people pages, most pronouns refer to the person about whom the wikipedia page is, whereas for other people mentioned in the page, Wikipedia
rarely refers them by pronouns, since other people in that page need not be mentioned frequently, as the page is not about them. Given this observation, including
gender in our model helps improving the performance in coreference
resolution part in our model. \\
\section{Conclusion}
As per our knowledge, this is the first time that this topic was
attempted. Although there exist Entity linking systems, which identify entities in a given text, but there aren't any entity linking systems that detect pronouns as entities and link them to a Wikipedia page. Use of such a Entity linking system in Information retrieval and summarization should improve the overall working of these applications. Our contribution includes a clean dataset consisting of 50
samples of Wikipedia pages from Wikipedia about people, and our models.\\
We manually annotated 50 samples, and cross-annotated them for
evaluation purpose. Our data covers a wide range of people, from
modern to ancient, and covers multiple professions. \\
 Our model performs better than previous successful models
in NER, Coreference Resolution and Entity Linking combined. Our model
is an optimized version of the baseline model, and it is optimized and
tuned for this specific area, Wikipedia people pages. This model would
definitely help in creating good data sets for training purposes in
entity linking tasks and we look forward to improving this system
further. Our model could be used to generate cleaner training set for
future entity linking tasks. \\
Since we are
using rule based features to optimize the model, and these rules work only for the pages about people in Wikipedia search, these rules will no
longer hold in another domain, so the model is not portable. But by
adding new rules, the model could possibly be extended to other
domains. 
\section{Future Work}
Based on our results and analysis, there are some things that can be
done in the future to improve the performance of the model:
\begin{itemize}
\item Collect more data. There are only 50 samples in our testing
  data set. This is mainly due to the time constraint of our
  project. Annotating Wikipedia pages is very time consuming. With
  more data in the testing data set, the performance of the models can
  be evaluated more thoroughly and fairly. It would help especially to
  include people from more areas. Currently, our samples include
  people from sports(mainly cricketers), chefs, scientist, and
  nobles. It doesn't make sense to extend the areas of people in our
  current data set, because if we include more areas, the number of
  people in each area would decrease, and the result may lose its
  statistical significance. So more data samples are necessary if the
  areas of people need to be extended. 
\item Use more information provided in the info box. For most
  Wikipedia people pages, there are information box available, which
  provides the birth date, first, middle and last name, professions, and
  etc. It would be beneficial to use more of these information in the
  model for identifying mentions. For example, in the Barack Obama
  information box, his full name, occupations, his spouse's name, his
  childrens' name, his predecessors' and successors' name and etc. are
  provided. These could all be incorporated into the model to help
  identify mentions, and possibly link mentions. 
\item Improve the rule-based features. It's possible to improve the
  rule-based features with linguistics. For example, adding a more
  comprehensive list of titles, or possible patterns of how people's
  names can be represented. 
\item Re-ranking Wikipedia search results. Currently in our model, we
  only take the topmost result returned from Wikipedia search API, thus
  we rely solely on Wikipedia's search algorithm for entity linking.  In
  the future, this part could be improved and instead of taking the topmost result, we can configure the model to take the top N results
  from Wikipedia search API, and re-rank them locally.Word embeddings are a meaningful way of representing entities, hence Word embedding
  could be used to rank the similarities between the entity and search
  results. we tried to use Word2Vec from Google, but it hard to find the names of people in its vocabulary list hence, the out of vocabulary ratio is too high for names of people. This is the reason why it was not useful.  A more suitable vocabulary is necessary if word embedding is
  to be used. 
\item Adding neural network. Based on the analysis, it is necessary
  and important to include context for entity linking. Convolutional
  neural network is known to be good at extracting contextual features. So, to further enhance the model 
  we can train a neural network so that we can extract contextual
  features and use those features along with the vector representation
  of an entity. The combination of the entity vector representation
  and contextual feature vector can be done in many ways, one way to
  do it is concatenation, to form a fixed size vector which represents
  an entity along with the contextual information. And we can use this
  vector representation for comparison with all wikipedia
  titles/articles. Doing this might improve the accuracy of the
  model. But a limitation, to this enhancement is the requirement of large training data set to train the neural network and training a neural network is computationally exhaustive and expensive.If a neural network is added to the model, significantly more
  training data is needed.   
\end{itemize}
\bibliographystyle{plainnat}
\bibliography{report}
\end{document}